\documentclass[10pt,twocolumn,letterpaper]{article}

\usepackage{cvpr}
\usepackage{times}
\usepackage{epsfig}
\usepackage{graphicx}
\usepackage{amsmath}
\usepackage{amssymb}
\usepackage{pdfpages}
\usepackage{caption}
\usepackage{subcaption}

\usepackage[pagebackref=true,breaklinks=true,letterpaper=true,colorlinks,bookmarks=false]{hyperref}

\cvprfinalcopy 
\usepackage[author={Devi}, open=true]{pdfcomment}

\newcommand{\argmax}{\mathop{\mathrm{argmax}}}

\newenvironment{packed_itemize}{
\begin{list}{\labelitemi}{\leftmargin=2em}
\vspace{-6pt}
  \setlength{\itemsep}{0pt}
  \setlength{\parskip}{0pt}
  \setlength{\parsep}{0pt}
}{\end{list}}
\def\cvprPaperID{467} 

\ifcvprfinal\fi
\begin{document}

\title{Cooperative Learning with Visual Attributes}

\author{Tanmay Batra$^{\ast}$\\
Carnegie Mellon University\\
{\tt\small tbatra@cmu.edu}
\and
Devi Parikh$^{}$\thanks{This work was done while T.B and D.P. were at Virginia Tech.}\\
Georgia Tech\\
{\tt\small parikh@gatech.edu}
}

\maketitle

\begin{abstract}
Learning paradigms involving varying levels of supervision have received a lot of interest within the computer vision and machine learning communities. The supervisory information is typically considered to come from a human supervisor -- a ``teacher'' figure. In this paper, we consider an alternate source of supervision -- a ``peer'' -- \ie a different machine. We introduce cooperative learning, where two agents trying to learn the same visual concepts, but in potentially different environments using different sources of data (sensors), communicate their current knowledge of these concepts to each other. Given the distinct sources of data in both agents, the mode of communication between the two agents is not obvious. We propose the use of visual attributes -- semantic mid-level visual properties such as furry, wooden, \etc~-- as the mode of communication between the agents. Our experiments in three domains -- objects, scenes, and animals -- demonstrate that our proposed cooperative learning approach improves the performance of both agents as compared to their performance if they were to learn in isolation. Our approach is particularly applicable in scenarios where privacy, security and/or bandwidth constraints restrict the amount and type of information the two agents can exchange.

\end{abstract}

\vspace{-10pt}
\section{Introduction}

\begin{figure}[t]
		\begin{subfigure}[b]{\columnwidth}
                \includegraphics[width=\columnwidth]{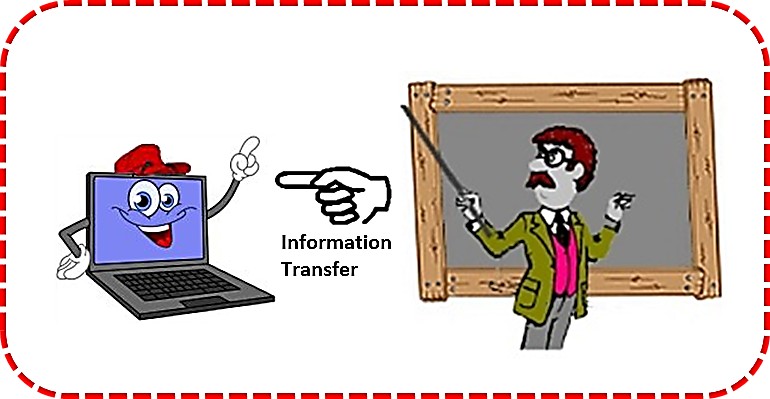}
                \caption{\textbf{Typically:} Humans are teachers training a machine to learn visual concepts.}
                \label{fig:teaser_typical}
        \end{subfigure}%
        ~\\
        \begin{subfigure}[b]{\columnwidth}
                \includegraphics[width=\columnwidth]{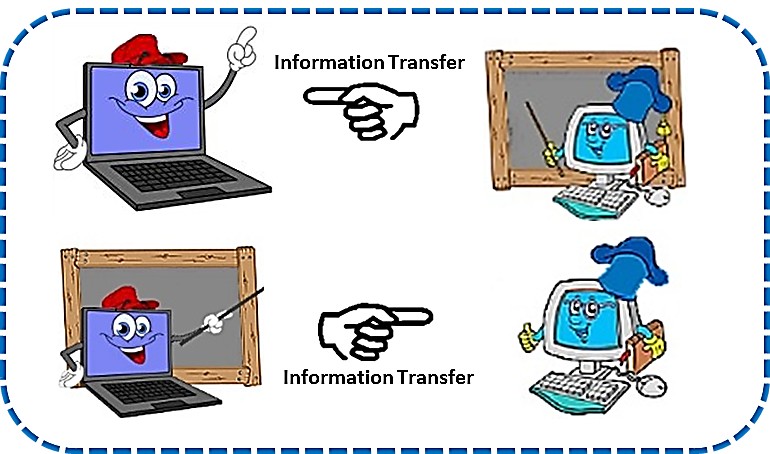}
                \caption{\textbf{Proposed:} Cooperative learning where two machines trying to learn the same concepts -- from potentially different sources of data using different sensors -- share their knowledge with each other \ie learn from each other.}
                \label{fig:teaser_cl}
        \end{subfigure}%
\caption{Our key idea: Cooperative Learning (bottom) as compared to traditional learning paradigms (top).}
\vspace{-10pt}
\end{figure}

Several learning paradigms exist that involve varying levels of supervision: from supervised learning that requires fully annotated datasets, to active learning that involves a human-in-the-loop annotating informative samples chosen by the machine, to semi-supervised and unsupervised learning that require little to no supervision. Whatever the level of supervision, the supervisory information is typically considered to come from a human supervisor -- a ``teacher'' figure considered to be more knowledgeable than the learning agent (Fig.~\ref{fig:teaser_typical}).

But what about exchanging information among ``peers''? Consider students (in the same class or going to different schools or colleges) learning the same concepts. Once taught in class, each student is most likely left with an incomplete and different understanding of the material. Students can continue to learn by studying the material themselves (\eg at home) \emph{and} by discussing the subject with each other (\eg studying in groups). As a result, learning continues to take place even without a teacher around. 

Inspired by this, we propose Cooperative Learning. Our envisioned scenario is the following: two (or more) agents are trying to learn the same set of concepts (say visual categories such as table, chair, cup). But they may be learning in different environments (\eg one is in an office and the other is in a living room or both are in two different offices) and/or from disjoint sources of data\footnote{Agents may have different sensors due to cost considerations (a small subset of robots may have the expensive sensors, while many others may have the relatively inexpensive ones), or some robots may be operating indoors vs. outdoors and/or during the day vs. at night.} (\eg one has an RGB camera and the other has a depth sensor). We propose an approach that, in spite of the differences in their sources of data and choice of sensors, allows the two agents to exchange information with each other about what they have currently learned about the visual categories of interest (common between the two agents), in turn improving each others understanding of the concepts (Fig.~\ref{fig:teaser_cl}). Such capabilities will be crucial as artificial agents become ubiquitous. While each agent may be learning about our world within its own domain and environment, it can result in collective learning across these agents, resulting in a whole that is larger than the sum of its parts.

Clearly, for this communication to happen, there must be a language in common between the two agents. The sensory data or the category models themselves can not be shared between the agents because both are using different sensors and hence different feature representations. The feature representations can not be concatenated because there is no alignment between the two sources of data from which the agents are learning. It is not clear how a robot with a Lidar sensor (say) can take advantage of an annotated but unaligned and disjoint set of RGB images (or their features) taken by a robot in a different environment. Additionally, due to security, privacy and/or bandwidth concerns, it may not be feasible for the agents to exchange the sensory data (or the features\footnote{A lot of progress has been made on inverting features to visualize the original image~\cite{vondrick2013hoggles, DBLP:journals/corr/MahendranV14}.}). These concerns arise for a variety of applications such as robots operating in our homes (privacy, and perhaps bandwidth concerns), robots being used for defense or security purposes (security concerns), robots responding to disasters (bandwidth concerns), or agents in autonomous vehicles (bandwidth concerns).

We propose the use of semantic mid-level visual properties or attributes -- \eg wooden, furry, young -- as the modality for this communication. The advantage of using attributes is that the semantic nature of attributes allows human annotators to annotate the same concepts in both domains/sensors, providing the common language. It also makes the models learnt by the agents, and the biases of their respective domains interpretable. Both agents can learn a mapping from their low-level features to these visual attributes, and a second mapping from visual attributes to categories. While the former is sensor- or domain-specific and can not be exchanged between agents, the latter is domain-independent and can be communicated between the two agents. For instance, one agent can communicate to the other agent that chairs tend to be either plastic or wooden, and the other might communicate to the first that chairs tend to have four legs. \footnote{We assume that while sharing imagery (of our homes say) across agents (across different homes) may have privacy concerns, this high-level information that describes categories in terms of their attributes is not sensitive. Moreover, as we will see later, these attribute-category relationship models tend to be very compact, alleviating bandwidth concerns.} This improves the understanding of both agents in terms of what chairs look like.

Concretely, our setup is the following: both agents are initialized with a few labeled examples in their respective domains. They both have access to (disjoint) sets of unlabeled data in their own domains. Both agents are learning in a semi-supervised manner. They use the labeled data to build category models in the low-level feature space (feature-category models). They then evaluate these models on the unlabeled data and retrieve images where the models are most confident of the classification label. These images and their predicted labels are added to the labeled set, and the feature-category models are updated. This process repeats for several iterations. This is the standard semi-supervised learning set up. Such approaches are known to suffer from semantic drift~\cite{Curran07minimisingsemantic}, where the concept that the agent is learning begins to morph into a semantically distinct concept as iterations go by. 

In our proposed cooperative learning setting, the small set of initial labeled images are annotated not just with the category labels, but also with attribute labels. In each iteration of semi-supervised learning, both agents learn attribute models in their respective domains (feature-attribute models) along with the feature-category models. They also learn a model of the interactions between categories and attributes. When retrieving the most confident images from the unlabeled pool to transfer to the labeled pool, the agents can use both the feature-category and attribute-category models. This is expected to improve performance. But more importantly, the attribute-category models, which tend to be quite compact, are communicated between the agents.
Both agents update their own attribute-category models using the newly received information from the other agent. We show that using this updated model to retrieve unlabeled images to transfer to the labeled pool results in further improvements in performance and resistance to semantic drift, without requiring additional human involvement.

\textbf{Contributions:}
Our main contribution is the introduction of a learning paradigm called cooperative learning, where agents that are trying to learn the same concepts share their current (potentially incomplete and different) models of the concepts with each other and accelerate each others learning. We propose using attributes as the common representation to communicate this information across agents. This allows the agents to communicate with each other even if each agent is learning from a different source of data using different sensors, even under privacy, security and/or bandwidth constraints. We believe that this approach may be among the very few that is applicable to this practical setting.
We show that cooperative learning leads to improvement in performance when compared to the performance of each agent learning independently. We show results on a synthetic dataset with varying amounts of noise and three standard real datasets of animals (Animals with Attributes~\cite{Lampert09learningto}), scenes (SUN~\cite{Patterson_sunattribute}), and objects (RGBD Objects~\cite{Lai_alarge-scale}).


\section{Related Work}
\textbf{Attributes:}
Visual attributes have been used quite extensively for a variety of tasks such as image classification~\cite{Kumar09attributeand,Lampert09learningto,Parkash:2012:ACF:2403072.2403100}, deeper image understanding~\cite{Farhadi09describingobjects}, image search~\cite{10.1109/CVPR.2012.6248026,CAVE_0287}, segmentation~\cite{Zheng_2014} and semi-supervised learning~\cite{shrivastava_eccv12} . They are mid-level semantic (\ie nameable) visual properties such as furry, wooden, young, \etc. They have been shown to be an effective modality for humans and machines to communicate with each other~\cite{10.1109/CVPR.2012.6248026,Parkash:2012:ACF:2403072.2403100,CAVE_0287,Lad2014}. We build upon these recent developments and propose attributes as a mode of communication among machines that may otherwise be unable to communicate due to differences in sources of data and sensors. The fact that attributes are semantic ensures that humans annotating the first few labeled images in both domains are annotating the same semantic concept. This provides a common ground between both domains, necessary for communication. 

\textbf{Semi-supervised learning:}
Among the various levels of supervision that have been explored in the community, semi-supervised learning approaches~\cite{Zhu05} try to achieve a good balance between maximizing accuracy while minimizing human input. A commonly used semi-supervised technique is the bootstrap or self-learning approach~\cite{Zhu05} where an agent initially learns from a small amount of labeled data. It then retrieves images from a large unlabeled pool whose labels it is most confident of, and transfers these to the labeled set to re-learn its model. Such approaches often suffer from semantic drift~\cite{Curran07minimisingsemantic}. Recently, visual attributes have been successfully used to resist this drift~\cite{shrivastava_eccv12}. This approach however relies on a human annotating all attribute-category relationships offline. In our approach, the machines communicate these attribute-category relationships to each other, alleviating the need for human involvement. Moreover, the attribute-category relationships estimated by machines are soft and more robust. Human annotated attribute-category relationships tend to be hard~\cite{shrivastava_eccv12,Lampert09learningto}, ignoring instance-level variations within categories.

\textbf{Active learning:} Visual attributes have been used as a medium for humans to communicate their domain knowledge to machines during an active learning loop for image classification~\cite{Parkash:2012:ACF:2403072.2403100,conf/cvpr/BiswasP13} or semi-supervised clustering~\cite{Lad2014}.  Another variant of active learning called coactive learning exists where the requirement that humans provides optimal labels to examples is loosened~\cite{conf/icml/ShivaswamyJ12}. Again, our approach eliminates the need for human involvement in each iteration by having machines communicate their current understanding with each other.

\textbf{Domain adaptation:}
Domain adaptation and transfer learning techniques have been applied to scenarios where a single agent learns visual categories (\eg monitor, keyboard) from one domain (\eg Amazon product images) but would like to apply its models to a different domain (\eg webcam images)~\cite{Saenko:2010:AVC:1888089.1888106,Hoffman:2012:DLD:2403006.2403059,6618936,journals/corr/abs-1301-3224}. Our setting is different. The two agents learning the visual categories will be using their models in their own domains. There is no ``adaptation'' required. Instead, the goal is for the two agents to communicate their current knowledge with each other so as to make them both more accurate in their \emph{own respective domains}. The agents continue to operate -- i.e., be used or ``tested'' -- in their own environments and domains. We assume that while the source of data and choice of sensors is different in both domains, the semantic distribution of categories -- \ie the attribute-category relationships -- are similar in both domains (\eg zebras in both domains are striped and have four legs). This is a reasonable assumption in many realistic scenarios (\eg robots in home environments). If one agent was learning about bears in a region where brown bears are common, and the other was learning about bears where white polar bears are common, this assumption would be violated. Finally, domain adaptation approaches typically rely on large quantities of labeled data from one of the domains (the source domain). In our setting, both domains use only a few labeled images.

\begin{figure*}[t]
\includegraphics[width =\textwidth ]{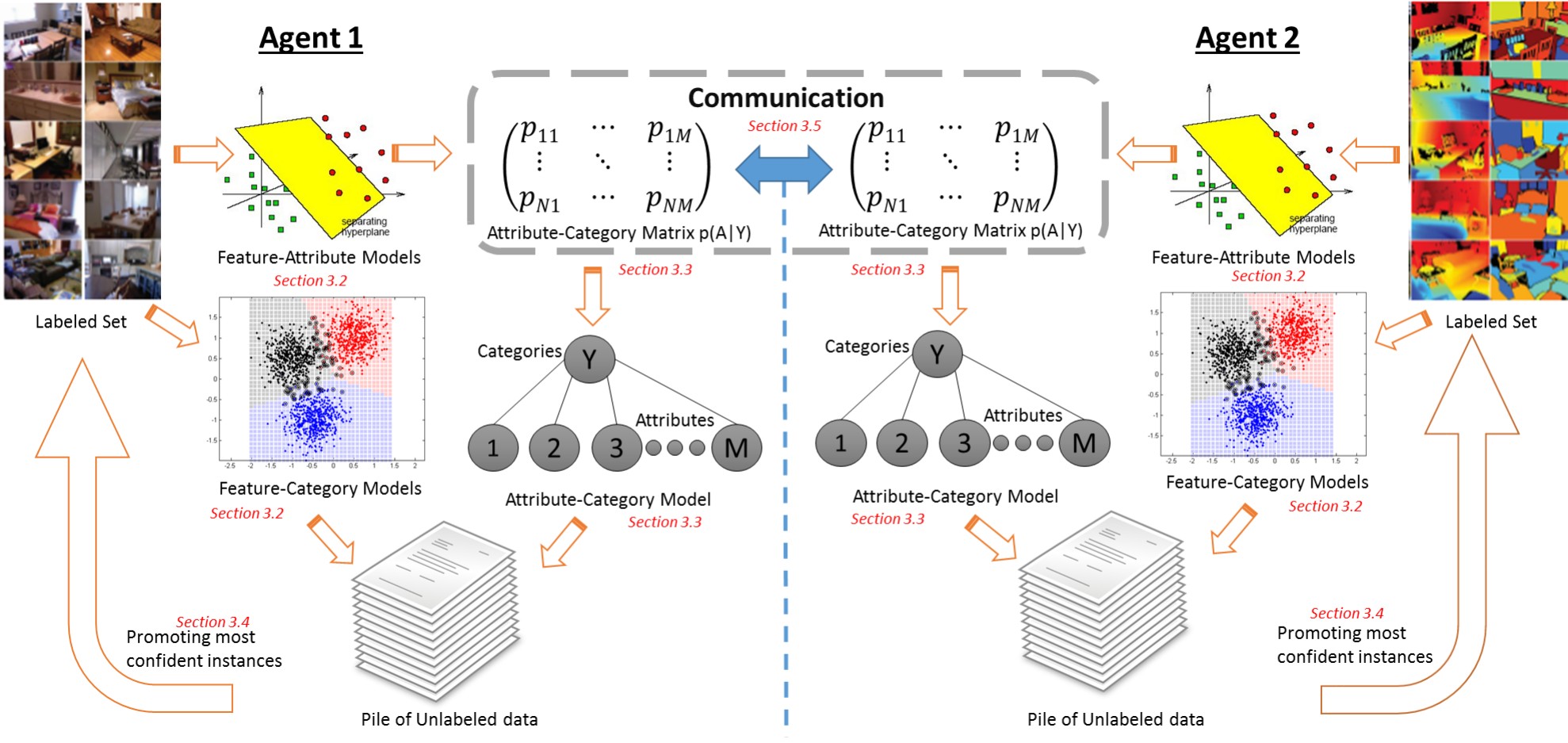}
\vspace{-20pt}
\caption{Overview of our approach}
\vspace{-10pt}
\label{fig:pipeline}
\end{figure*}

\textbf{Learning with multiple agents:}
Many approaches related to learning with multiple agents exist. One of them is called coactive learning~\cite{grecu98coactive}\footnote{Different from coactive learning in~\cite{conf/icml/ShivaswamyJ12}.}. It involves developing a distributed learning system to learn on large labeled datasets over multiple networks. They have a single domain and multiple `coactors' learning from that domain in parallel. The learning dataset is divided among the coactors. The communication is through transfer of misclassified training instances between the coactors. The focus of their approach is to parallelize the learning for a gain in speed. Our setting is different in that the different agents are learning in different environments and are communicating with the goal of improving accuracy. Collaborative machine learning leverages multiple sources of data (from a community of users)  for information retrieval~\cite{Marlin04collaborativefiltering} or recommendation tasks~\cite{Lee01collaborativelearning}.  In Siamese Neural Networks \cite{ZagoruykoCVPR2015,1467314}, multiple layers of neurons share weights whereas our approach involves multiple agents sharing category models in a semantic mid-level feature (attribute)space. Multi-task learning~\cite{journals/jmlr/Romera-ParedesABP12,xue_multi-task_2007} aims to optimize performance on multiple tasks, such as recognizing cats and dogs, by exploiting the relatedness of the tasks. In our setting, the tasks are the same, but the sources of data are different. Multi-view learning~\cite{Blum:1998:CLU:279943.279962} involves using different views (\eg features) for a single task. But these views are aligned \eg two different features extracted from the same image. Similarly in co-training or co-adaptation~\cite{Levin:2003:UIV:946247.946615, Christoudias:2006:CAS:1180995.1181013}, there are two views of the same data. A model is trained for each view. Each model identifies unlabeled images that it is most confident of. These images are fed to the other model as new training data annotated with the labels predicted by the first model. Again, this is feasible because the data is assumed to be aligned, just as in multi-view learning. Recent work on multi-modal learning \cite{Gupta_2016_CVPR, Hoffman_CVPR2016, castrejon2016learning, ayatar2016crossmodal} using deep networks also assumes that the data is aligned. In our setting, there is no alignment in the sources of data being used by both agents to learn. Finally, ``co-robots''~\cite{Goodrich:2007:HIS:1348099.1348100} refer to robots that are being trained to work along side humans. Our setting involves interactions between two machines without human involvement. A high-level description of ideas related to cooperative learning (without a discussion of modes of communication, algorithms, or implementation) is presented in~\cite{6295755}.


\section{Approach}

\subsection{Overview}
Fig.~\ref{fig:pipeline} shows an overview of our approach. Each part of this approach is described in the following subsections. Sec.~\ref{sec:init} describes the feature-category and feature-attribute models learnt by each agent. Sec.~\ref{sec:gm} describes the attribute-category models built by each agent. The multi-view approach used by each agent to transfer images from the unlabeled pool to the labeled pool is described in Sec.~\ref{sec:transfer}. Finally, details of the communication between the two agents are given in Sec.~\ref{sec:communication}.

\subsection{Feature-Category \& Feature-Attribute Models}
\label{sec:init}
We consider $K$ agents ($K$ = 2 in our experiments) learning in different domains in a semi-supervised bootstrap fashion. Each agent has a labeled set, an unlabeled set, and a test set (for measuring accuracy). The labeled set has category and attribute labels annotated. We train category classifiers (linear SVMs with Platt scaling) using the labeled images. These provide us with a probability $p_\text{FC}(y_i|x)$ indicating the probability that an image (feature vector) $x$ belongs to category $y_i$. This is our feature-category model which will be updated at each iteration as images are transferred from the unlabeled pool to the labeled set (Sec.~\ref{sec:transfer}). We also train binary attribute classifiers (linear SVMs with Platt scaling) for all $M$ attributes using the images in the labeled set. $p(a_m|x)$ is the probabilistic output of these classifiers indicating the probability of attribute $a_m$ being present in image $x$. These form our feature-attribute models and will also be updated at each iteration.

\subsection{Attribute-Category Model}
\label{sec:gm} 
Our attribute-category models consist of two parts: an attribute-category matrix and a probabilistic model that assigns an image to a category based on its attributes. We first describe the attribute-category matrix. We build a $M$ (number of attributes) $\times$ $N$ (number of categories) probability matrix $p(A|Y)$ representing attribute-category relationships. The entry $p_{ij}$ in this matrix stores $p(a_j|y_i)$ -- the fraction of images (in the labeled set) in the $i$-th category that have the $j$-th attribute present. 
This matrix, $p(A|Y)$, will be communicated from one agent to another (Sec.~\ref{sec:communication}). This matrix gets updated every iteration as new images are transferred to the labeled set.

The probabilistic model provides us with $p_\text{AC}(y_i|x)$ using the attribute-category matrix described above and the feature-attribute models described in Sec.~\ref{sec:init}. We formulate our probabilistic model as a Conditional Random Field (CRF) that is star-shaped: Y connected to all $M$ attributes (see Fig.~\ref{fig:pipeline}). Given an image $x$, the unary potentials of each attribute are $p(a_m|x)$ from the feature-attribute models. The edge potential connecting attribute $a_m$ to $y_i$ is $p(a_m|y_i)$ from the attribute-category matrix described above. We perform sum-product inference on this CRF to compute the marginal 

\vspace{-20pt}
\begin{equation}
\label{eq:gm}
p_\text{AC}(y_i|x) = \frac{1}{z}\prod\limits_{m=1}^M p_{m}(y_i|x)
\end{equation}

for a test image $x$. Each message, one for each attribute, $p_{m}(y_i|x)$ is computed as follows.

\vspace{-15pt}
\begin{align}
p_{m}(y_i|x) & = \sum\limits_{a_m\in\{0,1\}} p(y_i,a_{m}|x) \\
& = \sum\limits_{a_m\in\{0,1\}} p(y_i|a_{m},x)p(a_{m}|x) 
\end{align}

We assume that categories are fully specified by the visual attributes, \ie given the attributes, categories are independent of the image. This gives us

\vspace{-15pt}
\begin{align}
p_{m}(y_i|x) & = \sum\limits_{a_m\in\{0,1\}} p(y_i|a_{m})p(a_{m}|x) \\
& =\sum\limits_{a_m\in\{0,1\}} \frac{p(a_{m}|y_i)p(y_i)}{p(a_{m})}p(a_{m}|x)
\label{equation:gm}
\end{align}

Similar to ~\cite{Lampert09learningto}, we set the values of the attribute priors $p(a_{m}) = 0.5$ and category priors $p(y_i) = \frac{1}{N}$. 
Given a new image $x$ from the unlabeled set, we can now calculate $p_\text{AC}(y_i|x)$ for each category $i=1,2,....,N$. 

\subsection{Transferring Images}
\label{sec:transfer}
We move on to the next part in the pipeline: transferring images from a large unlabeled pool to the labeled set. Recall that for each agent, we have two category models that likely contain complementary information because the models represent two separate views. In order to reliably transfer images at every iteration, we use a multi-view framework. We combine two views: the feature-category models (Sec.~\ref{sec:init}) and the probabilistic attribute-category model (Sec.~\ref{sec:gm}). The combined probability of an image $x$ belonging to category $y_i$ is computed as

\vspace{-10pt}
\begin{equation}
p_\text{com}(y_{i}|x) = \frac{p_\text{FC}(y_{i}|x) + p_\text{AC}(y_{i}|x)}{2}
\end{equation}

We then calculate the entropy of this distribution, $E = -\sum_{i=1}^N p_\text{com}(y_{i}|x)\log(p_\text{com}(y_{i}|x))$. The images with lowest entropy values are selected for transfer. We transfer 2 images per iteration per (predicted) category (results with transferring 5 images are in the appendix).

Each image is ``annotated'' with $\argmax_{y_{i}} p_\text{com}(y_{i}|x)$ when transferred to the labeled set. Note that this same multi-view classification approach is used to classify images in the test set. To determine the attribute label of an unlabeled image transferred to the labeled set, we use the attribute-category matrix $p(A|Y)$ described in Sec.~\ref{sec:gm}. If the image has been transferred to the $i$-th category then the $j$-th attribute is annotated as 1 if $p_{ij}>0.5$ and 0 otherwise.

Finally, to further avoid semantic drift, we introspect the labeled set every few iterations. Similar to~\cite{shrivastava_eccv12}, after every 5 iterations, we use the multi-view framework to calculate the entropy of all images in labeled set, and prune images that are the least confident in each category. We prune 6 images from each category (same setting as in~\cite{shrivastava_eccv12}).

\subsection{Communication}
\label{sec:communication}
In our proposed cooperative learning approach, agents share their current knowledge about the visual categories with each other using visual attributes. In particular, they share the attribute-category matrix $p(A|Y)$. 
Since the agents have access to a few examples, they may learn incomplete and complementary information. Sharing this information is likely to supplement their knowledge and lead to improved performance and resistance to semantic drift.

Each agent updates its matrix to be the average of its current matrix and the matrix communicated by the other agent. \footnote{Other robust estimators can also be considered. In our experiments, averaging sufficed. Note that as iterations proceed, outliers become less likely.}
Let $p_{\text{agent-}k}(A|Y)$ be the current category-attribute matrix of the $k$-th agent. The updated matrix for agent $k$ after communication is

\vspace{-15pt}
\begin{equation}
\label{eq:com}
\hat{p}_{\text{agent-}k}(A|Y) = \frac{\sum_{k'} p_{\text{agent-}k'}(A|Y)}{K}.
\end{equation}
 
This updated matrix $\hat{p}_{\text{agent-}k}(A|Y)$ will be used in the subsequent iteration in the multi-view approach to transferring images to the labeled set (Sec.~\ref{sec:transfer}). Once images are transferred, the feature-category, feature-attribute and attribute-category models are all updated, and the iterations continue. Clearly, our approach is general enough to accommodate multiple agents. In our experiments, we use $K=2$ agents.

 
\section{Experimental Setup}
\label{sec:setup}

\textbf{Datasets:} We present our experimental results on one synthetic dataset and three existing real datasets. The synthetic dataset helps demonstrate our idea and allows us to control the amount of complementary information between the two agents.

The three real datasets that we experiment with are: Animals with Attributes (AWA)~\cite{Lampert09learningto}, SUN scene attributes (SUN)~\cite{Patterson_sunattribute} and the RGB-D Objects dataset~\cite{Lai_alarge-scale}. These datasets capture a wide array of categories (animals, indoor and outdoor scenes, household objects, \etc) and attributes (parts, habitats, shapes, materials, \etc). Both categories and attributes are annotated in these datasets. 

\textbf{Categories and Attributes:} For each of the datasets we select 10 categories at random. We use 10 attributes\footnote{~\cite{shrivastava_eccv12} found 10 attributes to be sufficient to constraint semi-supervised learning. More/fewer attributes/categories would make the downstream classification task more/less challenging. But that would affect both baselines and our approach. Our framework is a general one and not constrained to a fixed number of categories or attributes. The more attributes that are used, the more information the agents need to communicate with each other, which may run up against any bandwidth constraints.} from each dataset such that the attributes were present in a mix of the categories.\footnote{Note that this results in the agents communicating a $10 \times 10$ matrix to each other at each iteration. At double precision, this is ~0.0008 MB of data, as compared to ~2 MB or ~0.067 MB that would need to be communicated if images or DECAF features were exchanged. The latter, clearly, also have privacy concerns.}
Details on the features are provided below. A list of categories and attributes used for each dataset can be found in the appendix.

\textbf{Splits of data:} For each agent, we use disjoint sets of 5 images per category as the initial labeled set. The test set contains 200, 200 and 300 images per category from the AWA, SUN and RGB-D datasets, respectively. The remaining images from each category are in the unlabeled pool. To make the learning task challenging, we add $\sim$10k randomly selected images from the remaining categories of the datasets (other than the 10 we use) as distractors. We split the unlabeled set also into two disjoint sets (at random). The first set is the first agent's source of data to learn from, and the second is the second's.

\textbf{Features:} In addition to different sources of data to learn from, our approach allows the agents to also use different sensors (or feature spaces). Our main dataset (RGB-D) does naturally provide two modalities: RGB and DEPTH. We extract Hierarchical Matching Pursuit (HMP) features~\cite{bo_cvpr13,bo_iser12,bo_nips11,sun_icra13} for each domain provided by the authors. The HMP extractor uses RGB images to compute features for the RGB modality and depth images with surface normals for the Depth modality. Due to lack of other such datasets in the community, we use different sensors on the AWA and SUN datasets by extracting different features from the images as proxies. For AWA, the first agent uses SIFT (bag-of-words) features, while the second agent uses COLOR histograms (both made available by the authors of the dataset~\cite{Lampert09learningto}). For SUN, the first agent uses GIST and the second uses COLOR histograms (available with the dataset~\cite{Patterson_sunattribute}). 

\textbf{Baselines:}
We compare our approach to two baselines. 
\textbf{SSLIndLearner}: The standard bootstrap approach which does not involve  attributes. The two agents learn independently (no communication). Each agent learns from an initial pool of images labeled with categories, trains SVMs for each category, retrieves most confident images from the unlabeled pool, adds them to the labeled pool, updates its SVMs, and iterations continue.
\textbf{MultiViewSSLIndLearner}: The agents still learn independently, but the labeled set of images are annotated with category as well as attribute labels. They can thus build attribute-category models along with the feature-category models, and transfer images using our multi-view approach (Sec.~\ref{sec:transfer}). This baseline is the same as our approach except it is missing our proposed communication mechanism between the agents (Sec.~\ref{sec:communication}) -- key to cooperative learning.
\textbf{Rationale}: Comparing the second baseline to the first one quantifies the benefit of using attributes as a second view to constraint semi-supervised learning. Comparing our approach to the second baseline quantifies the additional benefit of our proposed cooperative learning approach. The parameter settings, transfer mechanisms, \etc. are consistent across baselines and our approach.

\textbf{Evaluation:}
In order to evaluate our approach we use two metrics that we evaluate at each iteration:
(1) Accuracy: class-average accuracy on the test set
(2) Purity of the labeled set: the percentage of images in the accumulated labeled pool that have been annotated with the correct label.

\begin{figure}[t]
\centerline{\includegraphics[scale=0.38]{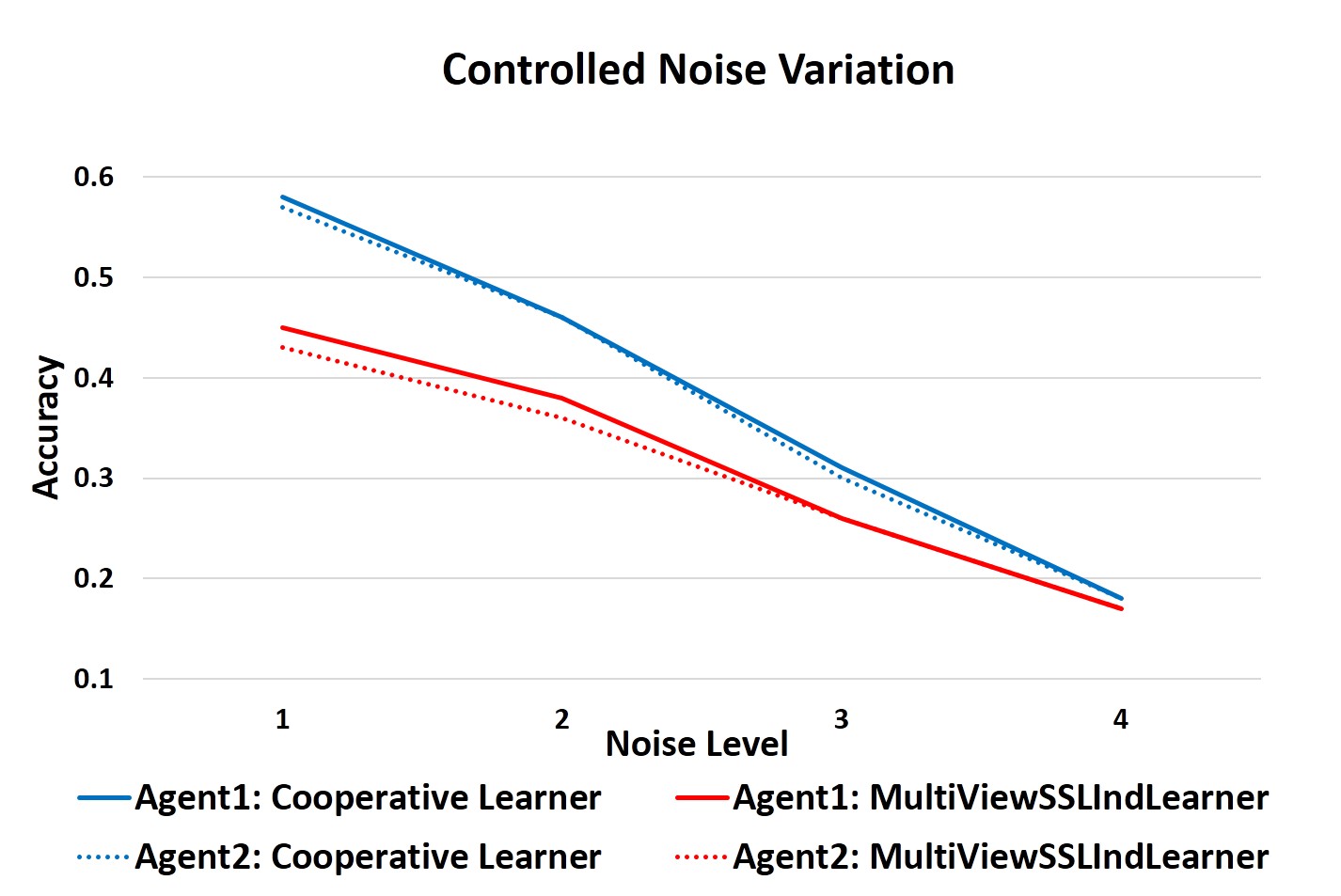}}
\vspace{-10pt}
\caption{Synthetic Dataset. On increasing the noise level, the ``complimentary'' nature of the two agents decreases and the margin of improvement between cooperative learning and the baseline reduces.}
\vspace{-10pt}
\label{res:noise}
\end{figure} 
\vspace{-2pt}
\section{Results and Discussions}

\begin{figure*}[t]
\centering
		\begin{subfigure}[b]{\columnwidth}
		\centerline{\includegraphics[scale=0.45]{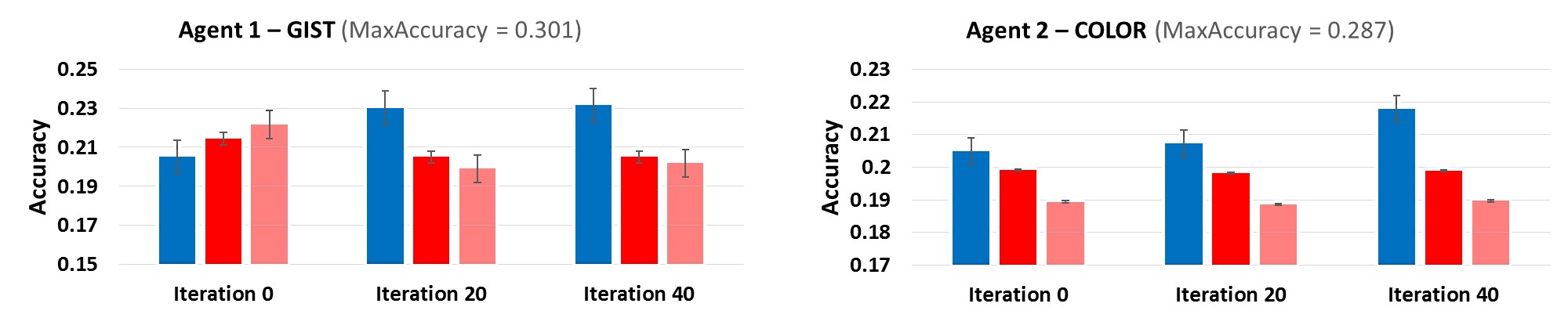}}
                \caption{SUN Dataset}
                \label{res:acc_sun}
        \end{subfigure}%
~\\
        \begin{subfigure}[b]{\textwidth}
                \centerline{\includegraphics[scale=0.45]{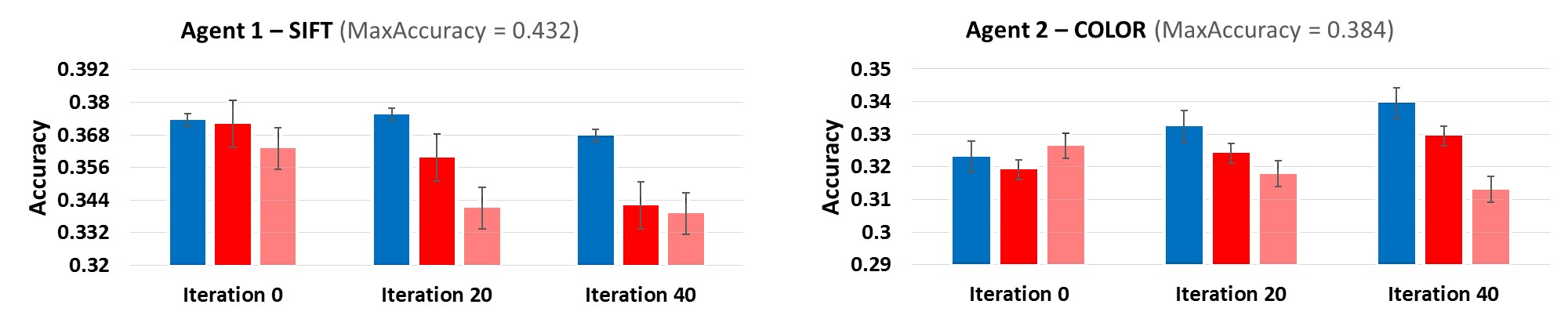}}
                \caption{AWA Dataset}
                \label{res:acc_awa}
        \end{subfigure}%
~\\
        \begin{subfigure}[b]{\textwidth}
                \centerline{\includegraphics[scale=0.45]{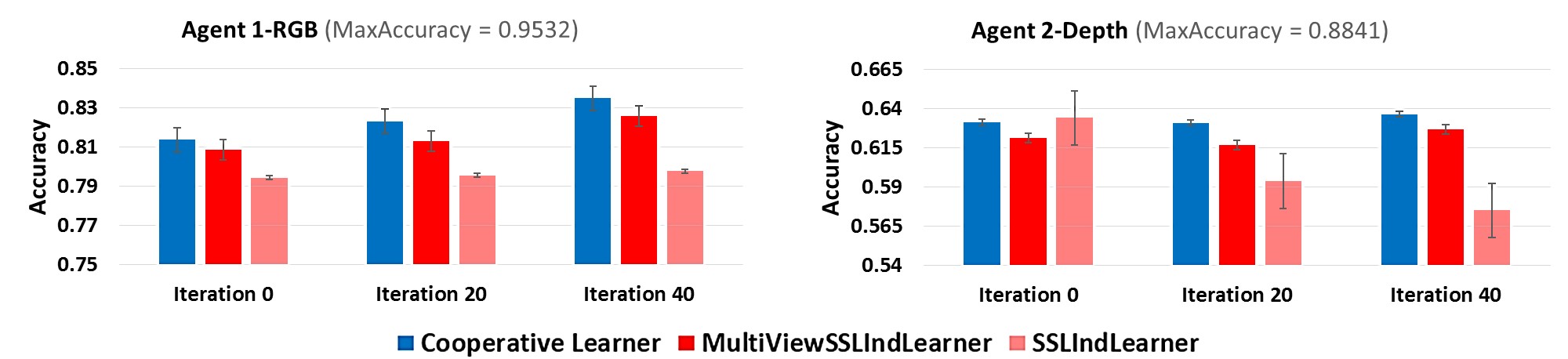}}
                \caption{RGB-D Dataset}
                \label{res:acc_rgbd}
        \end{subfigure}%
        \vspace{-5pt}
\caption{Accuracies of baselines compared to our approach for different datasets. We also mention MaxAccuracy for each agent. It is the accuracy of the agent (MultiViewSSLLearner) if trained in a fully supervised manner i.e. by annotating both the labeled set \emph{and} the unlabeled pool with category and attribute labels.}
        \vspace{-5pt}
\end{figure*}

\begin{figure*}[t]
\centering
		\begin{subfigure}[b]{\columnwidth}
		\centerline{\includegraphics[scale=0.45]{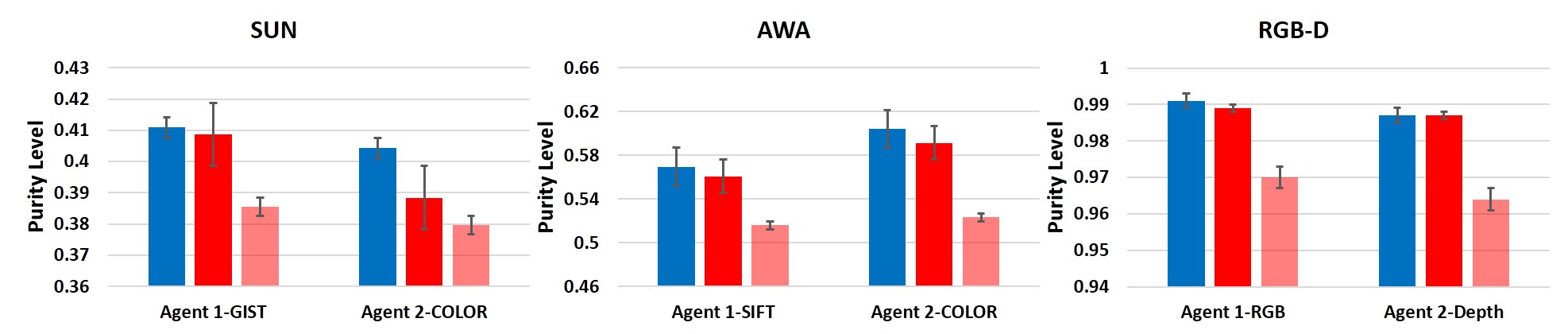}}
                \caption{Purity Level at Iteration 10.}
        \end{subfigure}%
~\\
        \begin{subfigure}[b]{\textwidth}
                \centerline{\includegraphics[scale=0.45]{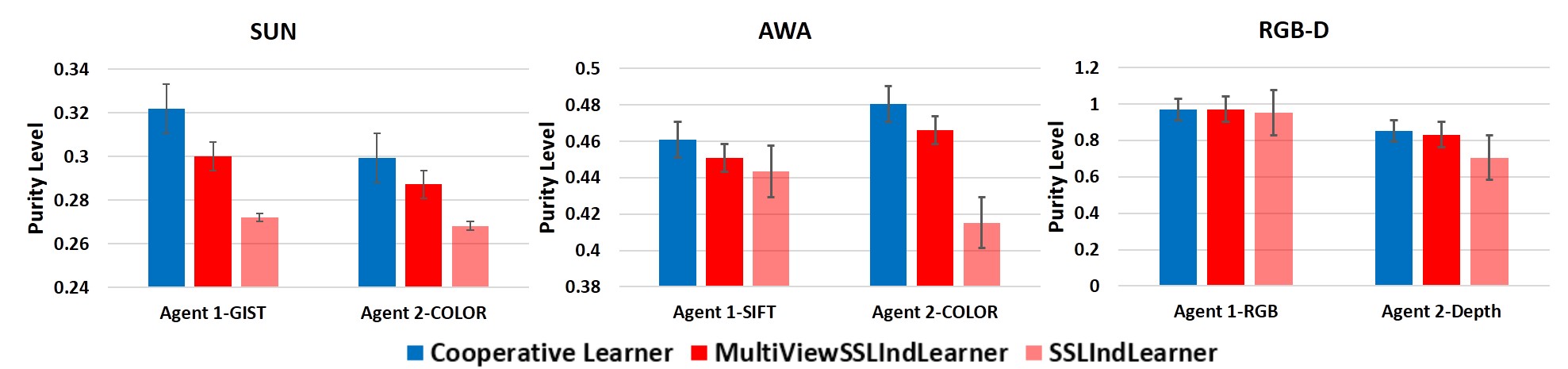}}
                \caption{Purity Level at Iteration 30.}
        \end{subfigure}%
        \vspace{-5pt}
\caption{Purity of the labeled set across iterations for different datasets. Comparing the top and bottom plots for the same dataset, we see that the purity decreases for all methods. 
Cooperative learning maintains higher purity across the board.}
\vspace{-10pt}
\label{res:purity}
\end{figure*}

\subsection{Synthetic dataset}
The contribution of our proposed approach over standard bootstrap semi-supervised learning is communication between the two agents via visual attributes. This is accomplished by averaging the attribute-category $p(A|Y)$ matrices of both agents as described in Eq.~\ref{eq:com}. We wanted to evaluate whether this average matrix tends to be more accurate at recognizing categories based on attributes, than the matrices of each agent individually. Furthermore, we wanted to evaluate how any improvement in accuracy is affected by how noisy the attribute predictions of the two agents are, and more importantly, by how complementary the attributes predictions of the two agents are.

To control the noise in the predictions, we take the ground truth attribute annotations of the images in the SUN dataset, and add zero-mean Gaussian noise to them with varying standard deviations.\footnote{All probabilities are followed by normalization to retain valid distributions.} These noisy annotations are treated as attribute predictions made by each agent.  Both agents estimate their $p(A|Y)$ from a labeled set of 50 images each. They use this matrix to classify images in a test set using Eq.~\ref{eq:gm}. This forms our baseline for this experiment. Our approach averages the two $p(A|Y)$ matrices and uses the resultant matrix to classify test images. Notice that the low-level features play no role here -- we use the attribute-category models alone (Sec.~\ref{sec:gm}). As a result, the MultiViewSSLIndLearner baseline uses just the attribute-category model.
When adding noise, we make half the attributes for an agent ``good'' and the other half ``bad'' (and \emph{vice versa} for the second agent). We start by setting our noise parameters such that for the given labeled data the good attribute classifiers have an accuracy of over $80\%$ and the bad ones vary between $50-65\%$ (chance is 50\%). We then leave the noise parameters the same for the bad attributes and increase the amount of noise added to the good ones, making them approach the bad ones, till all attributes are equally bad. As this happens, the complementary nature of the two agents starts decreasing. Results are shown in  Fig.~\ref{res:noise}. As expected, as the noise increases, the performance of both approaches decreases. But when the noise level is low for a subset of attributes (two agents are complementary), our cooperative learning approach results in significant improvements.

\vspace{-2pt}
\subsection{Real datasets}

We now present results on the three real datasets we experimented with, described in Sec.~\ref{sec:setup}. Fig.~\ref{res:acc_sun} shows results corresponding to the SUN dataset. 

Our approach outperforms both baselines. Moreover, unlike the baselines, our approach allows the agents to resist semantic drift, and increase performance over iterations. Also note that our stronger baseline MultiViewSSLIndLearner outperforms SSLIndLearner indicating that semantic attribute-based information helps constraint semi-supervised learning of low-level classifiers. Similar trends can be seen for AWA (Fig.~\ref{res:acc_awa}) and RGB-D (Fig.~\ref{res:acc_rgbd}) datasets.
For reference, also mentioned in Fig.s~\ref{res:acc_sun}, \ref{res:acc_awa} and \ref{res:acc_rgbd} is \emph{MaxAccuracy}. This corresponds to the upper bound accuracy of each agent (MultiViewSSLIndLearner) achieved by fully supervised learning. That is, this is the accuracy achieved by the agents when their entire unlabeled dataset was annotated with the ground truth category and attribute labels, and used to train their feature-category, feature-attribute and attribute-category models.
In our appendix, we have also shown how the accuracies of attribute models increase over iterations which shows that over time, the ``language'' of communication is also improving. We also compare our approach to a baseline of simple ensembling of two diverse models through averaging. See appendix for details.
We also evaluate our approach by measuring how pure the labeled set remains across iterations. We compute the percentage of images in the labeled set that have been assigned to the correct category. Fig.~\ref{res:purity} shows the trends. We can see that cooperative learning transfers images correctly more often than the baseline approaches.

\vspace{-2pt}
\subsection{Discussion}
In our approach, the attribute models are updated on-the-fly along with the category models. We also experimented with training attributes offline using annotations on a held out set of categories. We found this to be less effective. This may be because the attribute concepts learnt from a different set of categories do not generalize to the categories the agents are learning, and this bias is problematic to deal with when the agents have very few labeled images initially. 

In Eq.~\ref{eq:com}, all agents are weighted equally. We experimented with a weighted version where agents are weighted different based on the reliability of their learnt attribute classifiers. This did not help significantly, except in scenarios where the two agents have a high disparity in their accuracies. For instance, we experimented with using SIFT features for one agent and DECAF~\cite{Lampert09learningto} for the other (again, as a \emph{simulation} of different sensors) on the AWA dataset. The weighted version ensured that the first agent benefited a lot from the second agent without hurting the performance of the second agent (see appendix for results). This was because the weights essentially resulted in a unidirectional flow of information from the second agent to the first. Further exploration of mechanisms to estimate the expertise of each agent and modulate the communication based on that is part of future work.

\vspace{-6pt}
\section{Conclusions}

We propose a cooperative learning paradigm where different agents try to learn the same concepts, but in different environments using different sensors can communicate their current knowledge to each other. We propose the use of visual attributes as the mode of communication between the agents.  We explored this paradigm within the standard bootstrap semi-supervised learning loop. We found that our approach results in improved learning of both agents than if they were to learn independently, and both agents are able to resist semantic drift as compared to strong baselines. 

%
%

{\footnotesize
\bibliographystyle{ieee}
\bibliography{cl_cvpr}

\begin{thebibliography}{10}\itemsep=-1pt

\bibitem{ayatar2016crossmodal}
Y.~Aytar, L.~Castrejon, C.~Vondrick, H.~Pirsiavash, and A.~Torralba.
\newblock Cross-modal scene networks.
\newblock 2016.

\bibitem{conf/cvpr/BiswasP13}
A.~Biswas and D.~Parikh.
\newblock Simultaneous active learning of classifiers \& attributes via
  relative feedback.
\newblock In {\em CVPR}, pages 644--651. IEEE, 2013.

\bibitem{Blum:1998:CLU:279943.279962}
A.~Blum and T.~Mitchell.
\newblock Combining labeled and unlabeled data with co-training.
\newblock In {\em Proceedings of the Eleventh Annual Conference on
  Computational Learning Theory}, COLT' 98, pages 92--100, New York, NY, USA,
  1998. ACM.

\bibitem{bo_nips11}
L.~Bo, X.~Ren, and D.~Fox.
\newblock {Hierarchical Matching Pursuit for Image Classification: Architecture
  and Fast Algorithms}.
\newblock In {\em Advances in Neural Information Processing Systems}, December
  2011.

\bibitem{bo_iser12}
L.~Bo, X.~Ren, and D.~Fox.
\newblock {Unsupervised Feature Learning for RGB-D Based Object Recognition}.
\newblock In {\em ISER}, June 2012.

\bibitem{bo_cvpr13}
L.~Bo, X.~Ren, and D.~Fox.
\newblock Multipath sparse coding using hierarchical matching pursuit.
\newblock In {\em IEEE International Conference on Computer Vision and Pattern
  Recognition}, June 2013.

\bibitem{castrejon2016learning}
L.~Castrejon, Y.~Aytar, C.~Vondrick, H.~Pirsiavash, and A.~Torralba.
\newblock Learning aligned cross-modal representations from weakly aligned
  data.
\newblock In {\em Computer Vision and Pattern Recognition (CVPR), 2016 IEEE
  Conference on}. IEEE, 2016.

\bibitem{1467314}
S.~Chopra, R.~Hadsell, and Y.~LeCun.
\newblock Learning a similarity metric discriminatively, with application to
  face verification.
\newblock In {\em Computer Vision and Pattern Recognition, 2005. CVPR 2005.
  IEEE Computer Society Conference on}, volume~1, pages 539--546 vol. 1, June
  2005.

\bibitem{Christoudias:2006:CAS:1180995.1181013}
C.~M. Christoudias, K.~Saenko, L.-P. Morency, and T.~Darrell.
\newblock Co-adaptation of audio-visual speech and gesture classifiers.
\newblock In {\em Proceedings of the 8th International Conference on Multimodal
  Interfaces}, ICMI '06, pages 84--91, New York, NY, USA, 2006. ACM.

\bibitem{Curran07minimisingsemantic}
J.~R. Curran, T.~Murphy, and B.~Scholz.
\newblock Minimising semantic drift with mutual exclusion bootstrapping.
\newblock In {\em In Proceedings of the 10th Conference of the Pacific
  Association for Computational Linguistics}, pages 172--180, 2007.

\bibitem{6618936}
J.~Donahue, J.~Hoffman, E.~Rodner, K.~Saenko, and T.~Darrell.
\newblock Semi-supervised domain adaptation with instance constraints.
\newblock In {\em Computer Vision and Pattern Recognition (CVPR), 2013 IEEE
  Conference on}, pages 668--675, June 2013.

\bibitem{Farhadi09describingobjects}
A.~Farhadi, I.~Endres, D.~Hoiem, and D.~Forsyth.
\newblock Describing objects by their attributes.
\newblock In {\em Proceedings of the IEEE Computer Society Conference on
  Computer Vision and Pattern Recognition (CVPR}, 2009.

\bibitem{Goodrich:2007:HIS:1348099.1348100}
M.~A. Goodrich and A.~C. Schultz.
\newblock Human-robot interaction: A survey.
\newblock {\em Found. Trends Hum.-Comput. Interact.}, 1(3):203--275, Jan. 2007.

\bibitem{grecu98coactive}
D.~L. Grecu and L.~A. Becker.
\newblock {Coactive Learning for Distributed Data Mining}.
\newblock In {\em {Proceedings of the Fourth International Conference on
  Knowledge Discovery and Data Mining (KDD-98)}}, pages 209--213, New York, NY,
  August 1998.

\bibitem{Gupta_2016_CVPR}
S.~Gupta, J.~Hoffman, and J.~Malik.
\newblock Cross modal distillation for supervision transfer.
\newblock In {\em The IEEE Conference on Computer Vision and Pattern
  Recognition (CVPR)}, June 2016.

\bibitem{Hoffman_CVPR2016}
J.~Hoffman, S.~Gupta, and T.~Darrell.
\newblock Learning with side information through modality hallucination.
\newblock In {\em In Proc. Computer Vision and Pattern Recognition (CVPR)},
  2016.

\bibitem{Hoffman:2012:DLD:2403006.2403059}
J.~Hoffman, B.~Kulis, T.~Darrell, and K.~Saenko.
\newblock Discovering latent domains for multisource domain adaptation.
\newblock In {\em Proceedings of the 12th European Conference on Computer
  Vision - Volume Part II}, ECCV'12, pages 702--715, Berlin, Heidelberg, 2012.
  Springer-Verlag.

\bibitem{journals/corr/abs-1301-3224}
J.~Hoffman, E.~Rodner, J.~Donahue, K.~Saenko, and T.~Darrell.
\newblock Efficient learning of domain-invariant image representations.
\newblock {\em CoRR}, abs/1301.3224, 2013.

\bibitem{CAVE_0287}
N.~Kumar, P.~N. Belhumeur, and S.~K. Nayar.
\newblock {F}ace{T}racer: {A} {S}earch {E}ngine for {L}arge {C}ollections of
  {I}mages with {F}aces.
\newblock In {\em European Conference on Computer Vision (ECCV)}, pages
  340--353, Oct 2008.

\bibitem{Kumar09attributeand}
N.~Kumar, A.~C. Berg, P.~N. Belhumeur, and S.~K. Nayar.
\newblock Attribute and simile classifiers for face verification.
\newblock In {\em IEEE International Conference on Computer Vision (ICCV)},
  2009.

\bibitem{Lad2014}
S.~Lad and D.~Parikh.
\newblock {Interactively Guiding Semi-Supervised Clustering via Attribute-based
  Explanations}.
\newblock In {\em ECCV}, 2014.

\bibitem{Lai_alarge-scale}
K.~Lai, L.~Bo, X.~Ren, and D.~Fox.
\newblock A large-scale hierarchical multi-view rgb-d object dataset.
\newblock In {\em ICRA}, 2011.

\bibitem{Lampert09learningto}
C.~H. Lampert, H.~Nickisch, and S.~Harmeling.
\newblock Learning to detect unseen object classes by betweenclass attribute
  transfer.
\newblock In {\em CVPR}, 2009.

\bibitem{Lee01collaborativelearning}
W.~S. Lee.
\newblock Collaborative learning for recommender systems.
\newblock In {\em Proc. 18th International Conf. on Machine Learning}, pages
  314--321. Morgan Kaufmann, 2001.

\bibitem{Levin:2003:UIV:946247.946615}
A.~Levin, P.~Viola, and Y.~Freund.
\newblock Unsupervised improvement of visual detectors using co-training.
\newblock In {\em Proceedings of the Ninth IEEE International Conference on
  Computer Vision - Volume 2}, ICCV '03, pages 626--, Washington, DC, USA,
  2003. IEEE Computer Society.

\bibitem{DBLP:journals/corr/MahendranV14}
A.~Mahendran and A.~Vedaldi.
\newblock Understanding deep image representations by inverting them.
\newblock {\em CoRR}, abs/1412.0035, 2014.

\bibitem{Marlin04collaborativefiltering}
B.~Marlin.
\newblock Collaborative filtering: A machine learning perspective.
\newblock Technical report, 2004.

\bibitem{10.1109/CVPR.2012.6248026}
D.~Parikh, A.~Kovashka, and K.~Grauman.
\newblock Whittlesearch: Image search with relative attribute feedback.
\newblock {\em 2013 IEEE Conference on Computer Vision and Pattern
  Recognition}, 0:2973--2980, 2012.

\bibitem{Parkash:2012:ACF:2403072.2403100}
A.~Parkash and D.~Parikh.
\newblock Attributes for classifier feedback.
\newblock In {\em Proceedings of the 12th European Conference on Computer
  Vision - Volume Part III}, ECCV'12, pages 354--368, Berlin, Heidelberg, 2012.
  Springer-Verlag.

\bibitem{Patterson_sunattribute}
G.~Patterson and J.~Hays.
\newblock Sun attribute database: Discovering, annotating, and recognizing
  scene attributes.
\newblock In {\em CVPR}, 2012.

\bibitem{journals/jmlr/Romera-ParedesABP12}
B.~Romera-Paredes, A.~Argyriou, N.~Berthouze, and M.~Pontil.
\newblock Exploiting unrelated tasks in multi-task learning.
\newblock In N.~D. Lawrence and M.~Girolami, editors, {\em AISTATS}, volume~22
  of {\em JMLR Proceedings}, pages 951--959. JMLR.org, 2012.

\bibitem{Saenko:2010:AVC:1888089.1888106}
K.~Saenko, B.~Kulis, M.~Fritz, and T.~Darrell.
\newblock Adapting visual category models to new domains.
\newblock In {\em Proceedings of the 11th European Conference on Computer
  Vision: Part IV}, ECCV'10, pages 213--226, Berlin, Heidelberg, 2010.
  Springer-Verlag.

\bibitem{conf/icml/ShivaswamyJ12}
P.~Shivaswamy and T.~Joachims.
\newblock Online structured prediction via coactive learning.
\newblock In {\em ICML}. icml.cc / Omnipress, 2012.

\bibitem{shrivastava_eccv12}
A.~Shrivastava, S.~Singh, and A.~Gupta.
\newblock Constrained semi-supervised learning using attributes and comparative
  attributes.
\newblock In {\em European Conference on Computer Vision}, 2012.

\bibitem{sun_icra13}
Y.~Sun, L.~Bo, and D.~Fox.
\newblock {Attribute Based Object Identification}.
\newblock In {\em IEEE International Conference on on Robotics and Automation},
  2013.

\bibitem{6295755}
D.~Vidhate and P.~Kulkarni.
\newblock Cooperative machine learning with information fusion for dynamic
  decision making in diagnostic applications.
\newblock In {\em Advances in Mobile Network, Communication and its
  Applications (MNCAPPS), 2012 International Conference on}, pages 70--74, Aug
  2012.

\bibitem{vondrick2013hoggles}
C.~Vondrick, A.~Khosla, T.~Malisiewicz, and A.~Torralba.
\newblock {HOGgles: Visualizing Object Detection Features}.
\newblock {\em ICCV}, 2013.

\bibitem{xue_multi-task_2007}
Y.~Xue, X.~Liao, L.~Carin, and B.~Krishnapuram.
\newblock Multi-task learning for classification with dirichlet process priors.
\newblock {\em J. Mach. Learn. Res.}, 8:35--63, 2007.

\bibitem{ZagoruykoCVPR2015}
S.~Zagoruyko and N.~Komodakis.
\newblock Learning to compare image patches via convolutional neural networks.
\newblock In {\em Conference on Computer Vision and Pattern Recognition
  (CVPR)}, 2015.

\bibitem{Zheng_2014}
S.~Zheng, M.-M. Cheng, J.~Warrell, P.~Sturgess, V.~Vineet, C.~Rother, and P.~H.
  Torr.
\newblock Dense semantic image segmentation with objects and attributes.
\newblock {\em Proceedings of Computer Vision and Pattern Recognition (CVPR)},
  2014.

\bibitem{Zhu05}
X.~Zhu.
\newblock Semi-supervised learning literature survey.
\newblock Technical Report 1530, Computer Sciences, University of
  Wisconsin-Madison, 2005.

\end{thebibliography}
}

\section{Appendix}

\subsection{Transferring 5 images per iteration}

\begin{figure*}[t]
	\includegraphics[width =\textwidth]{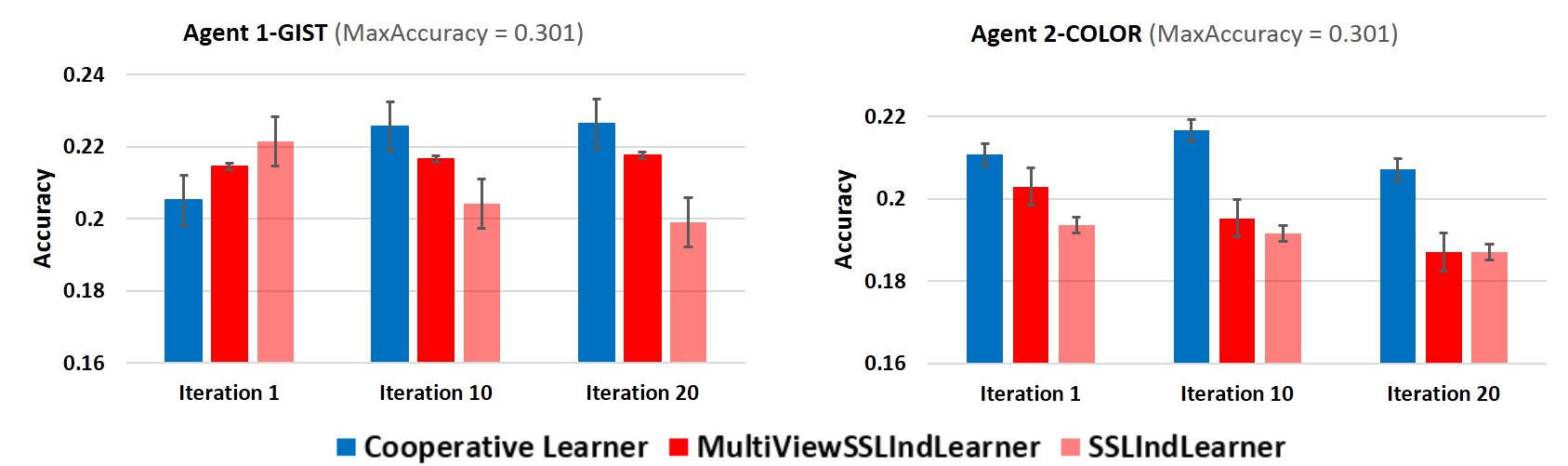}
	\caption{Accuracies of baselines compared to our approach (Cooperative Learner) for the SUN dataset when transferring 5 images per (predicted) category per iteration from the unlabeled pool to the labeled set (instead of 2 images in experiments in the main paper). MaxAccuracy is the accuracy of the agent (MultiViewSSLLearner) if trained in a fully supervised manner \ie by annotating both the labeled set and the unlabeled pool with category and attribute labels.}
	\label{res:sun}
\end{figure*}

We present results on the SUN dataset when transferring 5 images per category from the unlabeled pool to the labeled set (as opposed to 2 images per category as in the main paper). See Figure~\ref{res:sun}. Similar to the trends in the main paper, we find that accuracies are higher when using our proposed cooperative learning approach. However overall, we observe better performance when transferring 2 images. When transferring more images, there is a higher chance of images being assigned to the incorrect category when transferred, resulting in increased semantic drift.

\subsection{List of categories and attributes}

Following is the list of categories and attributes used in each of our datasets:

\textbf{SUN scene attributes (SUN)}:
\begin{packed_itemize}
\item Categories: We use categories from the second level of hierarchy. They are: shopping and dining; workplace (office building, factory, lab, etc.); home or hotel; transportation (vehicle interiors, stations, etc.); sports and leisure; cultural (art, education, religion, military, law, politics, etc.); water, ice, snow; mountains, hills, desert, sky; forest, field, jungle; man-made elements.
\item Attributes: trees; vegetation; foliage; natural light; natural; man-made; open area; enclosed area; no horizon; rugged scene.

\end{packed_itemize}

\textbf{Animals with Attributes (AWA)}:
\begin{packed_itemize}
\item Categories: chihuahua; rabbit; ox; mole; gorilla; giant panda; spider monkey; polar bear; elephant; sheep.
\item Attributes: white; gray; tough skin; bulbous; lean; paws; meat teeth; vegetation; forager; solitary.
\end{packed_itemize}

\textbf{RGB-D Objects dataset}:
\begin{packed_itemize}
\item Categories: food bag; food box; food can; food cup; garlic; hand towel; instant noodles; keyboard; kleenex; lemon.
\item Attributes: bag; cuboid; cylinder; ellipsoid; rectangle; fabric; metal; natural; paper; plastic.

\end{packed_itemize}

\subsection{Attribute models}

We evaluated learned attribute models on the SUN dataset and observed a statistically significant improvement in accuracies over 40 iterations. These accuracies are averaged across all the attributes. This shows that over time, the “language” of communication is also improving. 

Agent – 1 (Gist): Accuracy increases from 56.02\% to 58.10\% over 40 iterations.
Agent – 2 (Color): Accuracy increases from 59.15\% to 60.77\% over 40 iterations.
The standard error is 0.048\%.

\subsection{Baseline: Ensemble SSLIndLearner}

To ensure that our gain in accuracies is not coming just from simple ensembling of two diverse models through averaging the corresponding posterior probabilities of categories give images (similar to Eq(6) in main paper), we ran an additional baseline for the RGB-D dataset. We trained one SSLIndLearner in each domain using 5 labeled images per category, and then averaged the probability outputs for these two models and used that as the final probability of a category given a test image. Our cooperative learning approach beats this baseline. The accuracies of the baseline (and our cooperative learning approach in brackets) are:
Agent – 1 (RGB): 80.72\% (81.5\%) in iteration 1 and 83.001\%(83.71\%) in iteration 40.
Agent – 2 (Depth): 61.7\% (62.61\%) in iteration 1 and 62.22\% (63.9\%) in iteration 40.

\subsection{Non-uniform weighting of agents}

In our proposed cooperative learning approach, agents share their current knowledge about the visual categories with each other using visual attributes. Each agent constructs its own attribute-category matrix $p(A|Y)$ and shares it with other agents. Each agent updates its matrix to be the average of its current matrix and the matrix communicated by the other agent (Equation 7 in the main paper). We also experimented with a version where each agent is weighted differently when combining the matrices. The weights indicate which agent has a more reliable attribute classifier for each attribute. 

Specifically at each iteration, in addition to the attribute-category matrix $p(A|Y)$, each agent also shares a vector $Q$ of length $M$ with other agents which contains the agent's accuracies on all $M$ attributes. Each accuracy is estimated by running the attribute classifiers on the (original initialized) labeled set and comparing the predictions with the ground truth annotations. Now, the $i$-th column of the matrix $p(A|Y)$ is a vector $[p(a_i|y_1), p(a_i|y_2), ..., p(a_i|y_N)]$ where $y_1, y_2, ..., y_N$  are the categories. That is, the $i$-th column of the matrix $p(A|Y)$, say $p^i(A|Y)$, holds the probability of presence of attribute $a_i$ for all the categories. The matrix $p(A|Y)$ can be represented in terms of its column vectors as $[p^i(A|Y)]_{i=1}^M$. 

We now associate a scalar weight $w^i$ with the $i$-th column. Let $p_{\text{agent-}k}(A|Y) = [p_{\text{agent-}k}^i(A|Y)]_{i=1}^M$ be the current category-attribute matrix of the $k$-th agent and $Q_k$ be its classifier accuracies vector. The updated matrix for agent $k$ after communication is

\begin{equation}
\label{eq:com}
\hat{p}_{\text{agent-}k}(A|Y) = \sum_{k'} [w_{k'}^ip_{\text{agent-}k'}^i(A|Y)]_{i=1}^M
\end{equation}

For our experiment, we used two agents where weight $w_1^i$ is given as 

\begin{equation}
  w_1^i=\begin{cases}
    1, & \text{if $Q_1^i>Q_2^i$}.\\
    0, & \text{otherwise}.
  \end{cases}
\end{equation}

Similar equation holds for the weight $w_2^i$ corresponding to the second agent. Thus if the second agent is better than the first one for attribute $a_i$, the column $p_1^i(A|Y)$ in the first agent's matrix will be overwritten by the column $p_2^i(A|Y)$ in the second agent's matrix. This updated matrix $\hat{p}_{\text{agent-}k}(A|Y)$ will be used in the subsequent iteration in the multi-view approach for transferring images to the labeled set (Section 3.4 in main paper). Rest of the framework and experimental setup is same as described in the main paper. 

We experimented with the Animals With Attributes (AWA) dataset using SIFT features for one agent and DECAF features for the other. Attribute classifiers learnt in DECAF space are significantly more accurate than those learnt in SIFT space. We expect our weighted version to ensure that while the SIFT agent benefits from the DECAF agent, the DECAF agent is not hurt by the SIFT agent.

\begin{figure}[t]
\centerline{\includegraphics[scale=0.52]{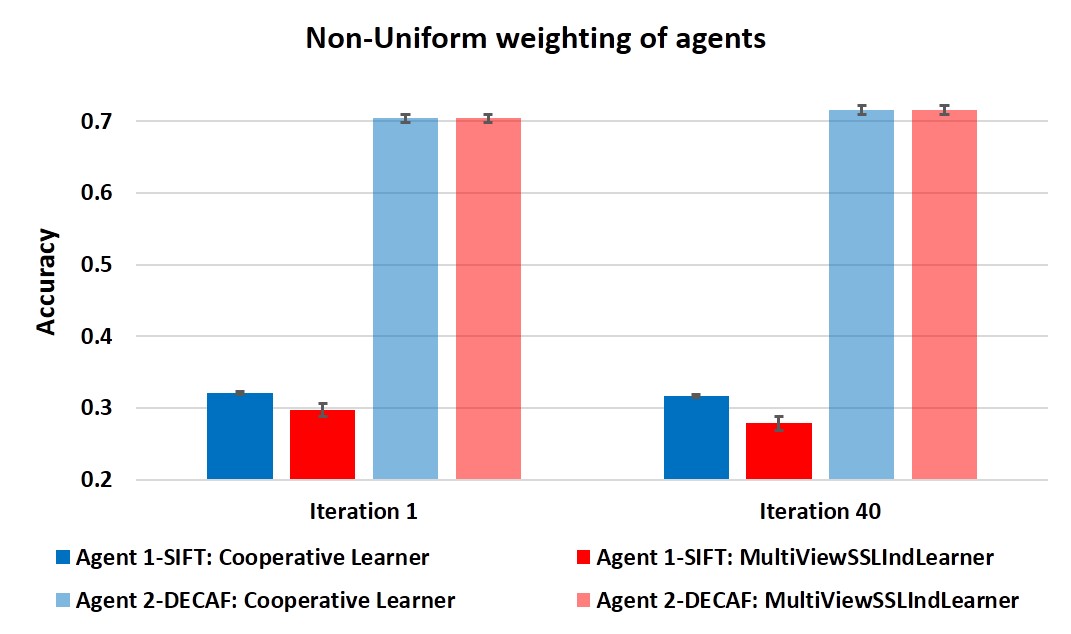}}
\caption{Accuracies of baselines compared to our (weighted) approach for both agents: SIFT and DECAF. In spite of SIFT performing much worse, it does not hurt the DECAF agent. Cooperative Learner and MultiViewSSLIndLearner for DECAF have the same performance because SIFT does not contribute any knowledge in our preliminary weighting scheme. The SIFT agent on the other hand benefits from the DECAF agent.}
\label{res:uni}
\end{figure}

The results are shown in Figure~\ref{res:uni}. We can observe that DECAF classifiers being better than SIFT classifiers for all 10 attributes, Agent-1 does not contribute anything to the knowledge of Agent-2 but the matrix $p(A|Y)$ of Agent-1 is overwritten by the matrix shared by Agent-2. As a result, the performance of the DECAF agent (Agent-2) is the same with or without cooperative learning. But the SIFT agent (Agent-1) benefits from cooperative learning. The weights resulted in a unidirectional flow of information from the second agent to the first. Further exploration of mechanisms to estimate the expertise of each agent and modulating the communication based on that is part of future work.

\end{document}